\documentclass{article}

\pdfoutput=1

\usepackage{PRIMEarxiv}

\usepackage[utf8]{inputenc} 
\usepackage[T1]{fontenc}    
\usepackage{hyperref}       
\usepackage{url}            
\usepackage{booktabs}       
\usepackage{amsfonts}       
\usepackage{nicefrac}       
\usepackage{microtype}      
\usepackage{lipsum}
\usepackage{fancyhdr}       
\usepackage{graphicx}       
\graphicspath{{media/}}     
\usepackage{times}
\usepackage{epsfig}
\usepackage{graphicx}
\usepackage{amsmath}
\usepackage{amssymb}
\usepackage{multirow}
\usepackage{subcaption}

\title{Zero-failure testing of binary classifiers}

\author{Ioannis Ivrissimtzis\\ 
Department of Computer Science, Durham University, UK\\ 
{\tt\small ioannis.ivrissimtzis@durham.ac.uk}
\and
Matthew Houliston\\
Serve Legal Ltd, London, UK\\
{\tt\small matthewhouliston@servelegal.co.uk} 
\and 
\\ Shauna Concannon\\ 
Department of Computer Science, Durham University, UK\\ 
{\tt\small shauna.j.concannon@durham.ac.uk}
\and
\\ Graham Roberts\\
Serve Legal Ltd, London, UK\\
{\tt\small grahamroberts@servelegal.co.uk}
}

\begin{document}
\maketitle

\begin{abstract}
We propose using performance metrics derived from zero-failure testing to assess binary classifiers. The principal characteristic of the proposed approach is the asymmetric treatment of the two types of error. In particular, we construct a test set consisting of positive and negative samples, set the operating point of the binary classifier at the lowest value that will result to correct classifications of all positive samples, and use the algorithm's success rate on the negative samples as a performance measure. A property of the proposed approach, setting it apart from other commonly used testing methods, is that it allows the construction of a series of tests of increasing difficulty, corresponding to a nested sequence of positive sample test sets. We illustrate the proposed method on the problem of age estimation for determining whether a subject is above a legal age threshold, a problem that exemplifies the asymmetry of the two types of error. Indeed, misclassifying an under-aged subject is a legal and regulatory issue, while misclassifications of people above the legal age is an efficiency issue primarily  concerning the commercial user of the age estimation system.
\end{abstract}



\section{Introduction}
\label{sec:sec1}

Binary decision problems are commonly approached as binary classification problems. There are two classes, the {\em positive} and the {\em negative}, and the task is to classify a sample as either positive or negative. Even though binary classification is the simplest form of a classification problem, assessing the performance of binary classifiers is not always straightforward. The main complication in the assessment of a binary classifiers emanates from the fact that there are two types of error. Positive samples misclassified as negative (Type I), and negative samples misclassified as positive (type II). Moreover, the costs associated with each error type could be unspecified, while user-defined costs can lead to opaque and rather arbitrary solutions. 

Our main motivation is to address this issue, aiming at developing system certification processes for vendors of commercial age estimation systems seeking regulatory approval for their products. In a typical scenario, we would like to offer such testing services to the developers of age estimation software that will be deployed at the self check-out tills in shops, aiming at identifying customers under a legal age threshold attempting to buy age restricted items. 

Clearly, this is a very asymmetric situation, not only regarding the costs associated with each error type, but also the stakeholders. A type I error means that a person under the legal age attempts to buy age restricted items, a situation that raises legal and regulatory issues. A type II error means that a person above the legal age will raise a false alarm at a self check-out till, meaning that they will have to be manually checked by a shop assistant, reducing the efficiency of the shop's operation. 

Our approach is to adapt to the asymmetry of the situation and assess performance with metrics derived from zero-failure testing. In particular, we require that the binary decision algorithm should first satisfy the regulator, demonstrating its reliability by passing a zero-failure test on a test set on positive samples. That is, all under-aged subjects in the test set should be correctly classified as positive samples. This is achieved by setting the operating point of the algorithm just above the highest age estimate of a positive sample. Next, having satisfied the regulatory requirements, we measure the performance of the algorithm by reporting its TNR, that is the probability that a subject above the legal age does not raise a false alarm, and thus, does not need to be manually checked. 

Zero-failure testing is a tool that is commonly used to assess the reliability of engineering products or biomedical products. In Section~\ref{sec:sec2}, we review the relevant theory of zero-failure testing, as well as the methods that are currently used to assess binary classification algorithms, and discuss differences and commonalities with our approach. The examples we use to illustrate the proposed method are organised as follows. 

In Section~\ref{sec:sec3-1} we present an example with synthetic data, simulating age estimates by the addition of some Gaussian noise to the actual age. We are experimenting with test sets comprising 60, 600 and 1500 positive samples in the 12-17 age range, each one corresponding to a different theoretical estimate for the reliability of the algorithm that passes the zero-failure test. The TNRs are computed either using an age threshold of 18, or a threshold of 25. The former is the usual TNR, the latter uses a hysteresis threshold, set here at 25, corresponding to the Challenge 25 policy operated in various shops, where false alarms by subjects under the age of 25 are considered as acceptable, if not desirable, behaviour of the algorithm.

In Section~\ref{sec:sec3-2} we compare two algorithms, CORAL-CNN \cite{cao2020} and OR-CNN \cite{niu2016} on the Morph2 database \cite{ricanek2006}. The four test sets of positive samples, again in the 12-17 age range, consist of 60, 200, 600, and 1550 images, and they are nested, that is, all smaller sets are subsets of the larger. The choice of nested test sets illustrates an interesting property of our approach. A zero-failure test on a set of positive samples will always be at least a strong as a zero-failure test on any of its subsets. This allows us to design a series of tests of increasing strength, where the test's strength does not depend on the algorithm we are testing. 

To the best of our knowledge, others types of testing do not produce such hierarchies of tests of increasing level of difficulty irrespective of the algorithm. Indeed, with the current approaches, there is no way to know with certainty and in advance if a performance metric will increase or decrease by the addition or removal of a sample from the test set; it all depends on how a certain algorithm will classify a certain sample. Of course, some databases are generally thought to provide more challenging test sets than others, but these are just empirical observations based on an adequate number of published results. 

Finally, in Section~\ref{sec:sec3-3} we apply the proposed zero-failure method  on the average human-estimated ages of the appa-real database \cite{agustsson2017}. The higher transparency of such human estimates allows several useful observations on how appropriate test sets for zero-failure testing should be constructed.

In summary, the contribution of the paper is a new method for testing binary classifiers and a demonstration of the method on the problem of age estimation for determining whether a subject is under a legal age threshold. The most interesting features of the proposed approach are: 
\begin{itemize}
    \item Addresses a possible asymmetry between the two types of error in a natural and parameter free way. First, the operating point is set such that a zero-failure test on a set positive samples is passed. Then, the TNR is reported as a measure of the efficiency that the system's user should expect. 
    \item Allows for the construction of a hierarchy of tests, the level of difficulty of which would always increase, independently of the tested algorithm. 
\end{itemize}

The rest of the paper is organised as follows. In Section~\ref{sec:sec2}, we discuss the background of zero-failure testing and of binary classification performance metrics. In Section~\ref{sec:sec3}, we illustrate the method with three examples and discuss its salient points, and we briefly conclude in Section~\ref{sec:sec4}. 

\section{Background}
\label{sec:sec2}

We first review zero-failure testing and then the metrics commonly used to assess binary classification algorithms. As a detailed review of the current state-of-the-art on age estimation is out of the scope of the paper, we refer the reader to the surveys \cite{fu2010,angulu2018,punyani2020,elkarazle2022}, which are also documenting the profound effect of deep learning on the field.

\subsection{Zero-failure testing}

Zero-failure testing is a well-established technique boasting strong mathematical foundations and a diverse range of applications. Its mathematical background is often studied under the problem of {\em success runs} \cite{philippou1986,fu1994,eryilmaz2018}, that is, the study of the likelihood of $n$ consecutive successes in the outcomes of an experiment, under certain probabilistic assumptions for the likelihood of a successful outcome. 

In more applied settings, zero-failure testing may appear in two closely related forms. In the first case, we want to compute probabilities for a device to fail within a certain time period, while our data do not contain any observations of failures \cite{bailey1997,bremerman2013}. In the second case, we want to design a demonstrator experiment, which, in the absence of any failures, will allow us to certify the system to a certain level of reliability, with a certain level of confidence. While our approach is more similar to the reliability demonstrator case, we note that its twin highlights a relevant for us property. Failures are rare events, and the zero-failure testing treats them as rare events. 

Mechanical or Electrical Engineering are typical application domains for zero-failure reliability demonstrators. In many cases, especially in the testing of some high-end, specialised products or processes, each individual run of the experiment can be very expensive and thus, experiments with a small number of runs are common. For example, \cite{webb2011} argues that 14 runs may suffice to demonstrate 90\% reliability with 90\% confidence, while in \cite{liang2023} the number of simulated experimental runs is similarly small to keep the simulation realistic. Thus, as one would expect, the number of experimental runs required to demonstrate a certain reliability with certain confidence is one of the key questions researchers focus on; see \cite{grundler2022} for a systematic discussion of zero-failure test design. 

In certain applications, failures are parameterised by the time they occurred in the duration of the test. Standard assumptions regarding their time distribution include the Weibull \cite{li2020a,li2020b,wang2024}, and the exponential distribution \cite{xu2014,yin2016}. When the test outcomes are not time parameterised, a simple approach is to assume equal probability of failure for all runs, in which case the number of failures follows the binomial distribution \cite{martz1979,han2015}. A well-established alternative is the Bayesian approach, starting from certain priors and updating them at each run of the test \cite{rahrouh2005,coolen2006}. Here, our test outcomes are not time parameterised, and for convenience, if needed, we will assume a binomial distribution for the number of failures. 

Finally, in the literature we reviewed there is no extensive discussion of an equivalent to an algorithm's operating point. Instead, one may assume that the parameters of the tested systems have already been calibrated by their developers with the aim of passing the zero-failure test. From this point of view, the difference with our approach is not the existence of an operating point, but rather that we choose it after the test. We note that this post-hoc choice could lead to optimistic reliability assessments compared to an equivalent test with predetermined operating point. Thus, as an alternative to the proposed method, one could require from the developers to specify the operating point of their algorithm in advance, however, that would also require a well-defined process on what happens next if the zero-failure test is not passed.


\subsection{Binary classification performance measures}

In the assessment of binary classifiers, it is quite common in practice to report success rates for the two classes separately: the True Positive Rate (TPR) and True Negative Rate (TNR), or the corresponding error rates: False Positive Rate (FPR), and False Negative Rate (FNR). Depending on the application, and how one wants to account for potential imbalances in the size of the two classes, one might report the Positive Predictive Value (PVV), that is, the true positives as a fraction of the total number of positive predictions, and the corresponding Negative Predictive Value (NPV). For direct performance comparisons based on a single scalar value, it is common to use combined metrics representing a form of average of the above, such as the F1-score, or the Matthews Correlation Coefficient (MCC). 

The above metrics require the operating point of the algorithm to be fixed. Other metrics, such as the Kolmogorov-Smirnov (KS) statistic and the Area Under the Curve (AUC) \cite{hand1997,fawcett2006}, measure algorithmic performance over all possible operating points. As they cannot be used to estimate performance on the specific operating point of a deployed algorithm, they do not always give a satisfactory answer to the problem of the relative costs between the two error types. Indeed, as \cite{hand2009} points out, the implicit assumptions of the KS statistic on these relative costs are unrealistic, while in the case of AUC, these costs are not independent, as they should be, of the algorithmic behaviour. As an alternative, in \cite{hand2009,hand2014} user defined relative costs in the form of beta distributions are proposed, however, as we mentioned above, in many settings user defined costs are not acceptable solutions. 

Finally, an approach to performance assessment with similarities to zero-failure testing, is to sample some points of interest on the ROC curve, reporting the algorithm's performance on the negative class for given FPRs. For example, \cite{yu2023} reports algorithmic performance for FPR values of 1\% and .1\%. In principle, by reporting algorithmic performance on the negative class for 0\% FPR, one applies zero-failure testing. However, performance at 0\% FPR is very rarely reported in the literature, and when it is reported, as for example in the Fingerprint Verification Competition of 2004 \cite{fvc2004}, it is only as part of a long, exhaustive list of metrics. As we will discuss later, we believe that the main reason for considering the algorithmic performance at 0\% FPR to be unrepresentative is because positive test sets are not designed for zero-failure testing.

\section{Zero-failure testing on age estimation}
\label{sec:sec3} 

\subsection{Test set size estimation - a synthetic data example}
\label{sec:sec3-1} 

In this first example, the positive samples are in the 12-17 age range, and the negative samples in the range 18-50. We avoid samples of very low or very high age, which in all likelihood will be classified correctly, inflating the measured performance. For simplicity, the actual age values are integers, but the Gaussian noise we add on them to simulate the error of the age estimation algorithm is real numbers. Again for simplicity, we have the same number of samples from each year under 18, and the same number of samples from each year from 18 and above. 

To compute a number of required positive samples, which will determine the power of the zero-failure test, we follow an empirical procedure assuming equal probabilities of failure across samples. Adopting the standard terminology, the {\em reliability} of the system is a statement on its behaviour, here it corresponds to the value of $1-p$, where $p$ is the probability of a failure. The {\em confidence} is a lower bound on the probability that the reliability statement is correct. Thus, for example, .90 reliability and .95 confidence means a .95 confidence that the probability of failure is less than .10. 

\begin{figure*}[ht]
\centering
\includegraphics[width=0.32\textwidth]{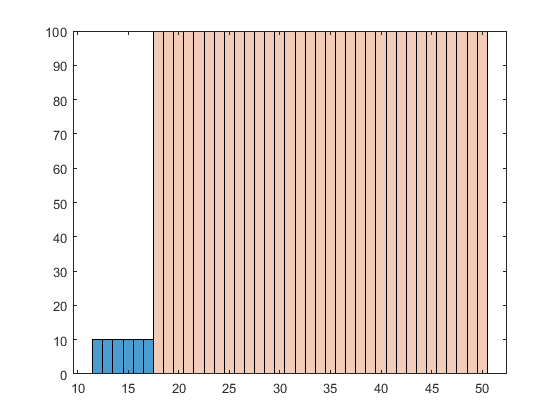} \hfill 
\includegraphics[width=0.32\textwidth]{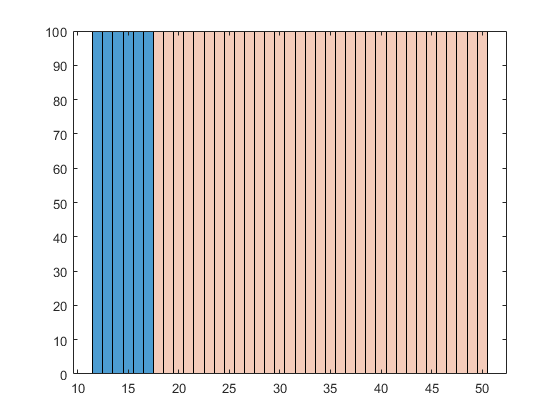} \hfill 
\includegraphics[width=0.32\textwidth]{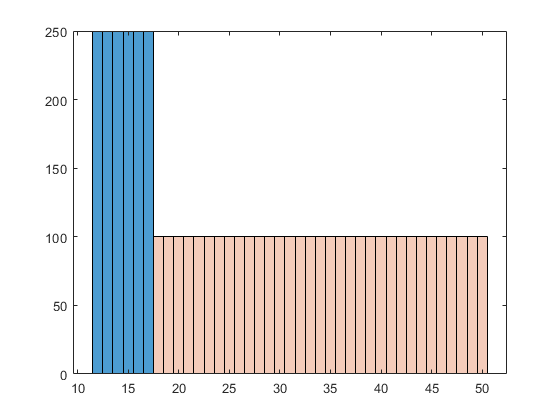} 
\includegraphics[width=0.32\textwidth]{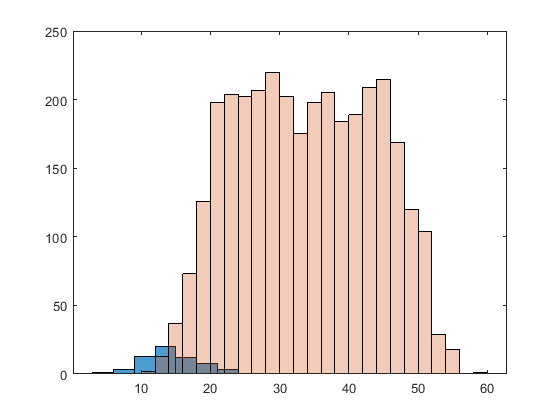} \hfill 
\includegraphics[width=0.32\textwidth]{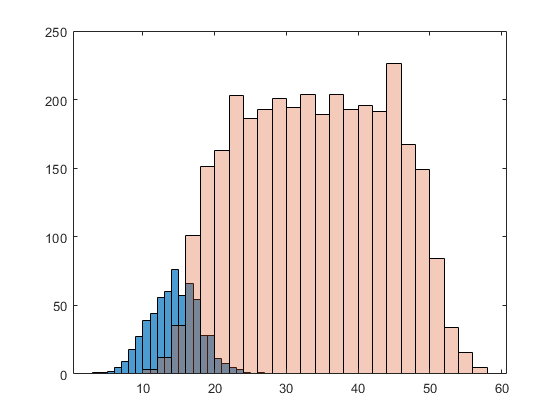} \hfill 
\includegraphics[width=0.32\textwidth]{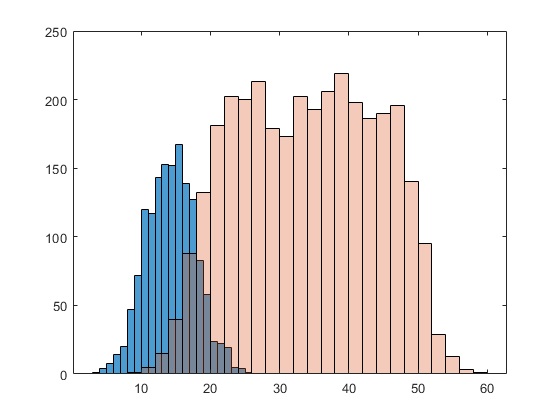} \hfill 
\caption{{\bf Top} Histograms of the actual ages of the subjects. The age range 12-17 is shown in blue, the range 18-50 in orange. {\bf Bottom:} Histograms of the estimated ages of the subjects. The estimates for the subjects in the age range 12-17 are shown in blue, the estimates for the subjects in the range 18-50 in orange. {\bf Left to right:} Confidence - reliability pairs of (.95, .95), (.95, .995), and (.95, .998).}
\label{fig:syntheticHist}
\end{figure*}

Assuming independent trials, simple computations on the binomial distribution show that the numbers of trials $N$ required to achieve confidence $c$ on reliability $1-p$ is 
\begin{equation}
N = \frac{\ln (1-c)}{\ln (1-p)},
\label{eq:setSize}
\end{equation}
see \cite{burr1979}, and \cite{kalegaonkar2017} for a simple exposition. We chose some confidence and reliability pairs that are often used in practice
$$
(c,1-p) = (.95, .95), (.95, .995), (.95, .998), 
$$ 
with the corresponding numbers of positive samples for the zero-failure tests being $N$ = 58.4, 597.6, and 1496.3, respectively, which we rounded them up to $N$ = 60, 600, and 1500. Thus, the number of samples from each year in the range 12-17 were 10, 100, and 250, respectively. Regarding the number of negative samples, we always use 100 samples for each year of age between 18 and 50, meaning that the TNRs are computed on a set of 3300 samples, a set size commensurable to what we would have expected in practice. Figure~\ref{fig:syntheticHist} shows the histograms of the actual and the estimated ages. 

Next, we find the highest age estimate corresponding to a positive sample and set it as the operating point of the binary classifier. We report the success rates on all negative samples $\textbf{TNR}_{18}$, and also success rates $\textbf{TNR}_{25}$ on negative samples in the range 25-50. The former is the usual TNR; the latter which uses the age of 25 as a hysteresis threshold corresponds to the Challenge 25 policy operated by certain sellers of age restricted item, where people deemed to be under the age of 25 are asked to produce an ID. The results are summarised in Table~\ref{tbl:synthetic}. 

\begin{table}[ht]
\begin{center}
\begin{tabular}{|c|c|c|c|c|} 
 \hline
            & $N$    & $threshold$ & $\text{TNR}_{18}$ & $\text{TNR}_{25}$ \\ \hline
(.95, .95)  & 60   & 23.4 & .821 & .975 \\ \hline
(.95, .995) & 600  & 26.9 & .713 & .895 \\ \hline
(.95, .998) & 1500 & 25.1 & .766 & .939 \\ \hline
\end{tabular}
\caption{The number of positive samples $N$, the lowest operating $threshold$ that successfully passes the zero-failure test, the usual TNR, and the TNR corresponding to a Challenge 25 policy.}  
\label{tbl:synthetic}
\end{center}
\end{table}

Looking at these zero-failure examples from the perspective of each stakeholder, the {\em regulator} endorses a zero-failure test, which, when successfully passed satisfies them on the reliability of the system. This zero-failure test is used to set the operating point of the binary decision algorithm. The system's {\em operator}, e.g. a seller of age restricted items, knows that after setting the operating point at the level that satisfies the regulator, they should expect an $\text{TNR}_{18}$ probability that a person over the legal age of 18 would not need to be manually checked, and an $\text{TNR}_{25}$ probability that a person over the age of 25 would not need to be manually checked. 

We emphasise here that we use the test size computation in Eq.~\ref{eq:setSize} as an empirical rule of thumb, without making any strong claims regarding reliability and confidence. We note that for the chosen error distribution, here a Gaussian with 0 mean and a standard deviation of 3, the assumption of equal probabilities of failure $p$ for all positive samples could have indeed been satisfied if all positive samples had the same actual age, let say the current mean value of 15.5. However, that would mean a very skewed distribution of the actual ages in the test set. Moreover, the property of equal probability of failure, which in that cases would have been imposed on this synthetic example, would not reflect the behaviour of actual age estimation algorithms. 

\subsubsection{Outliers}

It is an interesting and somewhat counter-intuitive observation on the previous example that the threshold for $N=600$ is higher than the threshold for $N=1500$. This is due to an outlier in the sample from the Gaussian distribution that was used to generate synthetic age estimates. Indeed, in close inspection, this outlier is visible in Figure~\ref{fig:syntheticHist} (bottom, middle). We believe that the possible existence of such outliers in a test dataset is perhaps the main reason behind the current practice of avoiding  zero-failure testing. Indeed, there are certain situations where one would like to prevent a single or a small number of outliers to penalise heavily the measured performance of a system. 

One standard approach to achieve this is to allow for one or a small number of failures, in our case to operate the algorithm at a point that will allow a certain number of failures \cite{martz1985,leemis1996,guo2012}. While allowing for a small number of failures is a legitimate approach, we believe that it does not suit well a setting like ours, where testing is inexpensive and can easily be repeated several times. Instead, we believe that zero-failure testing is the most transparent approach, as long as a high quality test set ensures that outlier estimates can be attributed to the behaviour of the algorithm, rather than to issues with the test set. 

Indeed, while in practice a certain number of failures, or most commonly a certain percentage of failures in the form of a FNR$>$0, is considered acceptable, the interpretation of these failures remains ambiguous. On the one hand, one can see the choice of an operating point at FNR$>$0 as introducing an application specific trade-off between the two error types, implying their relative costs. That is, we allow a few false negatives in order to reduce significantly the number of false positives. On the other hand, in practice, an FNR$>$0 is also an allowance for imperfections in the test set. These could either come from noisy, low quality samples, or correspond to some hard examples, the correct classification of which would go beyond what one would consider as a reasonable requirement. Zero-failure testing resolves this ambiguity in a transparent way; it separates these two issues which otherwise would be difficult to disentangle. In Section~\ref{sec:sec3-3}, we will further discuss the issue of exceptionally hard to classify samples in the test set.


\subsection{Hierarchies of nested zero-failure test sets - an algorithmic age estimation example}

In the three simulated zero-failure tests in Section~\ref{sec:sec3-1} we used three different random samples from the Gaussian distribution. This is the equivalent of assessing different algorithms with each test, and the algorithm we assessed on the $N=600$ test produced an outlier estimate. Nevertheless, we note that even if we had run all tests on a single set of synthetic data, there is still no guarantee that the zero-failure tests on the smaller test sets would result to lower operating thresholds. Indeed, it could still happen that the most challenging positive samples, the ones with the highest error, were in the smaller set. We note that this is an issue with any type of algorithmic testing; the computed values of the performance metrics cannot be decoupled from the properties of the test set. It is only after several researchers have used a dataset, and have published their experimental results, that emerges a consensus on how challenging compared to others a certain dataset is. 

However, in the case of zero-failure testing, there is a partial order of the level of difficulty of the test sets, defined not by the number of positive samples they contain, in which case it would have been a total order, but by the subset relation. Indeed, the inclusion of an additional sample in the zero-failure test set, can only increase the operating threshold, that is, it can only make the test harder. 

\begin{figure}[h]
    \centering
    \includegraphics[width=0.4\columnwidth]{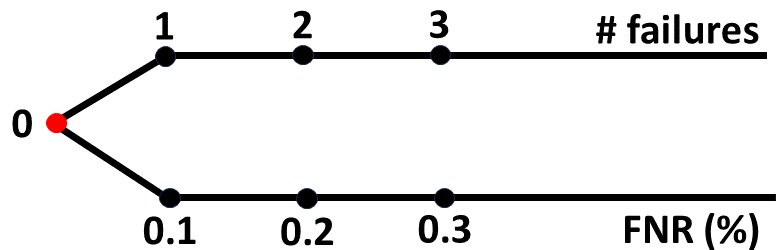}
    \caption{Zero-failure testing as the intersection of the family of tests that allow a fixed number of failures (top) and the family of tests that allow a fixed ratio of failures (bottom).}
    \label{fig:schematic}
\end{figure}

We note that this property is not unique to zero-failure testing but common to all testing methods allowing a fixed number of failures. In contrast, error metrics corresponding to a fixed operating point on the ROC curve do not have this property as they are based on a fixed ratio of failures rather than a fixed number of failures, but nevertheless, they are usually preferred as they scale naturally with the size of the dataset. Zero-failure testing is at the intersection of these two approaches, that is, the fixed number and the fixed ratio of failures, and shares their characteristics. See Figure~\ref{fig:schematic} for a schematic. 

\subsubsection{Example} 
\label{sec:sec3-2} 

In the second example, we apply zero failure testing to comparatively assess the COnsistent RAnk Logits (CORAL) CNN algorithm, and the Ordinal (OR) CNN algorithm, on the Morph2 database \cite{ricanek2006}. The age estimates outputted by the two algorithms were taken from the experimental logs published in github by the authors of \cite{cao2020}, who used OR-CNN \cite{niu2016} as a comparator for CORAL-CNN. 

The test set of the Morph2 database we used, comprised 11044 samples from subjects in the 0-54 age range. There were exactly 1550 samples from subjects in the age range 12-17 and we used them to construct the zero-failure test set {\em zFail-1550}. Then we sampled a random subset {\em zFail-600} with 600 samples, a further random subset {\em zFail-200} with 200 samples, and a further random subset {\em zFail-60} with 60 samples. Thus, our zero-failure test sets were ordered by the subset relationship 
\begin{equation}
\mbox{zFail-60} \subset \mbox{zFail-200} \subset \mbox{zFail-600} \subset \mbox{zFail-1550}. 
\end{equation}

For each algorithm the authors of \cite{cao2020} trained three different neural network models, each one using a different random seed to initialise the model's weights. We report TNRs for all three models, choosing each time the operating point as the lowest value for which the corresponding zero-failure test set is passed successfully. Table~\ref{tbl:tnr} shows TNRs computed on the 5268 samples of the Morph2 test set from subjects in the 18-49 age range; TNRs on the 2939 samples in the 25-49 age range, which correspond to the expected efficiency of the neural network models when operating a Challenge 25 policy; and finally, TNRs on the 1517 samples in the 30-49 range, which correspond to the expected efficiency of the models when operating a Challenge 30 policy. 
\begin{table*}[ht]
\begin{center}
\begin{tabular}{|c|c|c|c|c||c|c|c|} 
\cline{3-8}
            \multicolumn{2}{c|}{} & \multicolumn{3}{|c||}{CORAL-CNN} & \multicolumn{3}{|c|}{OR-CNN} \\ \cline{3-8}
            \multicolumn{2}{c|}{} & seed-0 & seed-1 & seed-2 & seed-0 & seed-1 & seed-2 \\ \hline
\multirow{4}{*}{Challenge 18} & zFail-60 & 0.3973 & 0.4056 & 0.4724 & 0.5065 & 0.5762 & {\bf 0.5870} \\ 
& zFail-200 & 0.3352 & 0.4056 & 0.4071 & 0.2808 & 0.4238 & {\bf 0.5304} \\ 
& zFail-600 & 0.3352 & 0.3492 & {\bf 0.3501} & 0.2808 & 0.3035 & 0.3251 \\ 
& zFail-1550 & 0.3352 & {\bf 0.3492} & 0.2509 & 0.0538 & 0.1316 & 0.2075
 \\ \hline 
\end{tabular}

 \vskip 0.1cm

\begin{tabular}{|c|c|c|c|c||c|c|c|} 
 \hline 

\multirow{4}{*}{Challenge 25} & zFail-60 & 0.6550 & 0.6710 & 0.7557 & 0.7799 & 0.8489 & {\bf 0.8574} \\ 
& zFail-200 & 0.5692 & 0.6710 & 0.6676 & 0.4794 & 0.6883 & {\bf 0.8074} \\ 
& zFail-600 & 0.5692 & {\bf 0.5941} & 0.5910 & 0.4794 &  0.5141 & 0.5478 \\ 
& zFail-1550 & 0.5692 & {\bf 0.5941} & 0.4365 & 0.0919 & 0.2293 & 0.3617
 \\ \hline 
\end{tabular}

 \vskip 0.1cm

\begin{tabular}{|c|c|c|c|c||c|c|c|} 
 \hline
\multirow{4}{*}{Challenge 30} & zFail-60 & 0.8827 & 0.9018 & 0.9380 & 0.9361 & 0.9558 & {\bf 0.9697} \\ 
& zFail-200 & 0.8312 & 0.9018 & 0.8952 & 0.7535 & 0.8991 & {\bf 0.9532} \\ 
& zFail-600 & 0.8312 & {\bf 0.8490} & 0.8471 & 0.7535 & 0.7924 & 0.8082 \\ 
& zFail-1550 & 0.8312 & {\bf 0.8490} & 0.7053 & 0.1727 & 0.4206 & 0.6144 \\ \hline 
\end{tabular}
\caption{{\bf Top to bottom:} TNRs computed on the 5268 samples in the Morph2 test set from subjects in the 18-49 age range (Challenge 18); on the 2939 samples in the 25-49 age range (Challenge 25); and the 1517 samples in the 30-49 age range (Challenge 30).}
\label{tbl:tnr}
\end{center}
\end{table*}

As a first observation, we note that the CORAL-CNN algorithm, which in \cite{cao2020} was shown to give lower Mean Absolute Error (MAE) and Root Mean Square Error (RMSE) rates than OR-CNN, outperforms OR-CNN on the zero-failure testing, provided that the zero-failure test is strong enough, that is, when the positive samples in the zero-failure test sets are 600 or 1550. One possible explanation is that the very end of the tail of the error distribution is heavier in the case of OR-CNN, and thus, affects the measured performance when the zero-failure tests get stronger and the probability of sampling from that very end of the tail increases. We also note that the comparative performance of the two algorithms does not seem to be greatly affected by the random seed. In fact, on the zFail-600 and zFail-1550 tests all three models of CORAL-CNN outperform all three models of OR-CNN. 

\begin{figure}[ht]
    \centering
    \includegraphics[width=0.4\columnwidth]{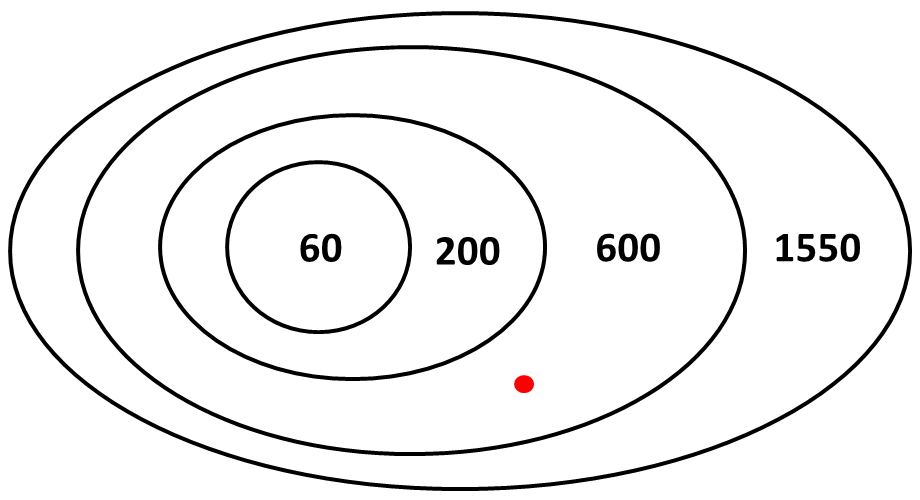}
    \caption{The nested test sets $\mbox{zFail-60} \subset \mbox{zFail-200} \subset \mbox{zFail-600} \subset \mbox{zFail-1550}$. In the CORAL-CNN, seed-1 classifier, the highest age estimate on zFail 1550, depicted with a red dot, comes from a sample in its subset zFail-600.}
    \label{fig:nested}
\end{figure}

In a second observation, we note that in all columns of Table~\ref{tbl:tnr} the TNRs appear, as expected, in decreasing but not necessarily strictly decreasing order. When two tests return the same TNR on the same model it means that the highest age estimate comes from a sample in the intersection of the two sets, that is, from a sample in the smallest set. For example, in CORAL-CNN, seed-1, the highest age estimate on zFail-200 comes from a sample in its subset zFail-60, while the highest age estimate on zFail 1550 comes from a sample in its subset zFail-600, see Figure~\ref{fig:nested} for a schematic. 

\subsection{Considerations for the design of zero-failure datasets - a human estimation example} 
\label{sec:sec3-3} 

Finally, we work an example with human age estimates of the apparent age of a subject, which are provided as part of the appa-real database \cite{agustsson2017}. Our aim is to use the higher transparency of the human voting system to illustrate good practices and potential pitfalls in the design and construction of a zero-failure test set. 

\begin{figure}[ht]
    \centering
    \includegraphics[width=0.66\textwidth]{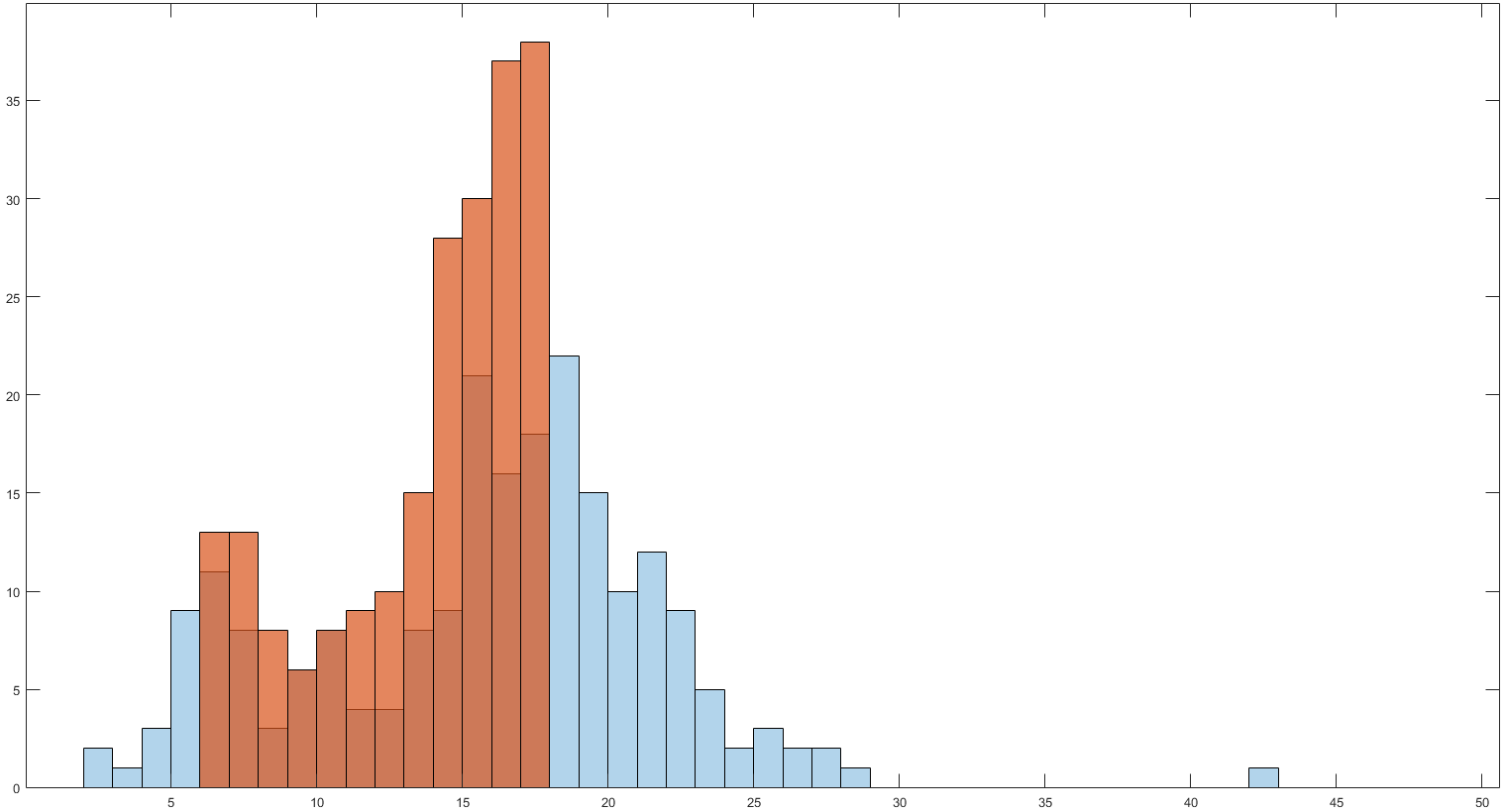}  
    \caption{Shown in red, the histogram of the actual ages of the 215 subject in the validation set of the appa-real database in the 6-17 age range. Shown in blue, the histogram of the average age estimates by human estimators.}
    \label{fig:appaReal-under18}
\end{figure}

The appa-real database consists of a {\em training}, {\em validation}, and a {\em test} dataset and we used the validation dataset which happened to be more suitable for illustrating our observations. Figure~\ref{fig:appaReal-under18} shows the histogram of the actual ages of the 215 subjects in the 6-17 age range in the validation set, and the histogram of the average human estimates. The age of each subject was estimated by about 30 different people. We note that the current release of the database also provides the single person estimates, and if we were to use these and take the worst case, the operating thresholds would have been higher. On the other hand, we also note that for a human, the problem of age estimation from a low resolution digital photo is much more challenging compared to age estimations in face-to-face encounters. 

\begin{figure*}
     \centering
     \begin{subfigure}[b]{0.29\textwidth}
         \centering
         \includegraphics[width=\textwidth]{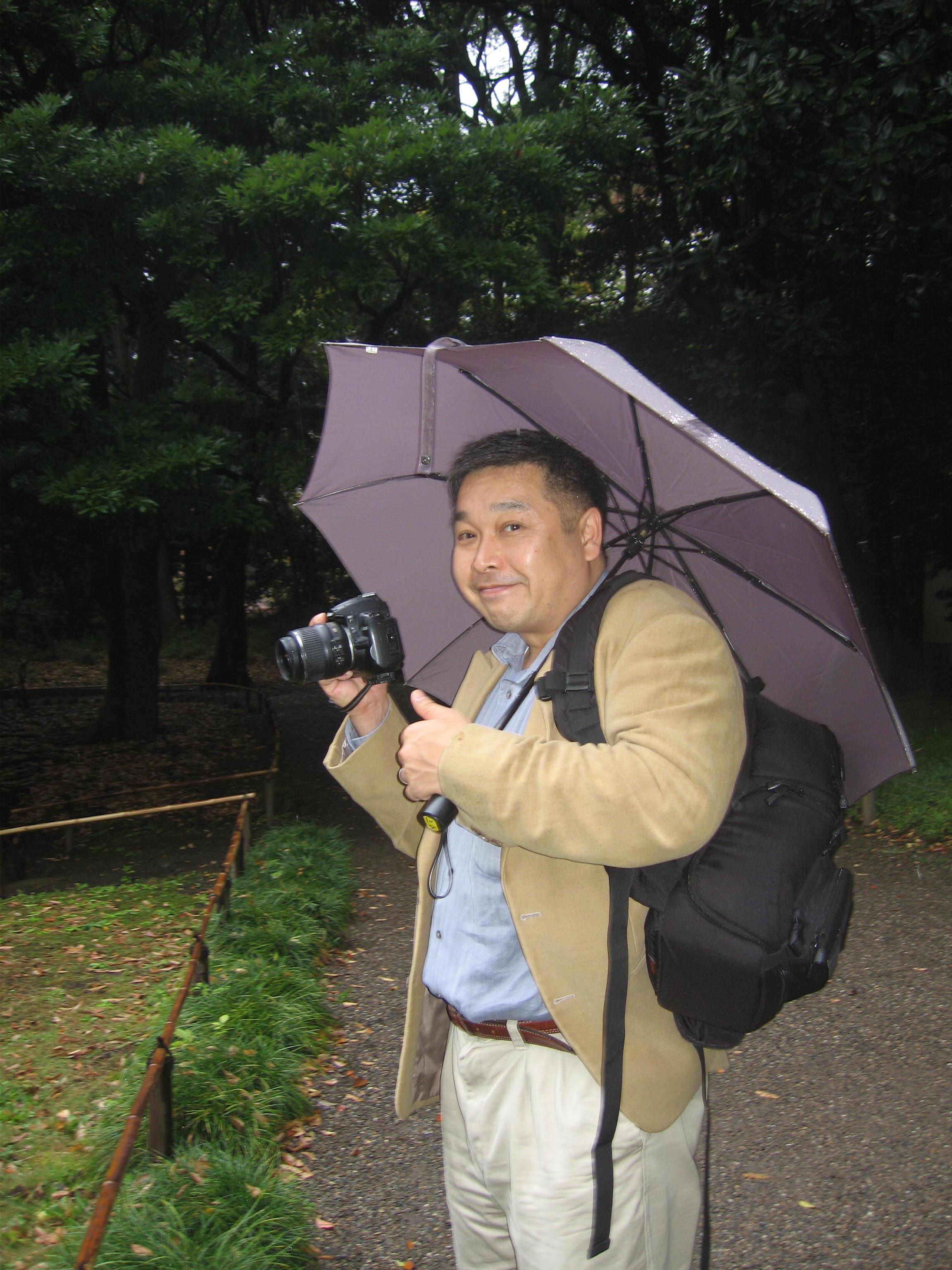}
         \caption{estimated 42.81 ({\bf actual 6})}
         \label{fig:subject1}
     \end{subfigure}
     \hfill
     \begin{subfigure}[b]{0.29\textwidth}
         \centering
         \includegraphics[width=0.89\textwidth]{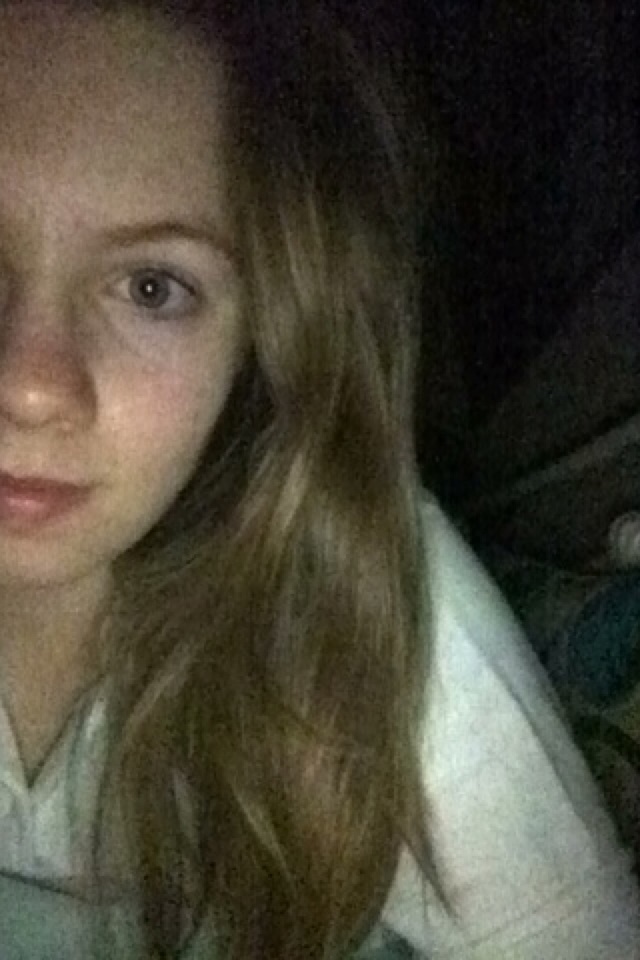}
         \caption{estimated 28.23 ({\bf actual 13})}
         \label{fig:subject2}
     \end{subfigure}
     \hfill
     \begin{subfigure}[b]{0.39\textwidth}
         \centering
         \includegraphics[width=\textwidth]{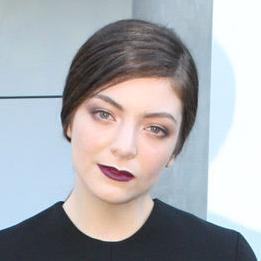}
         \caption{estimated 27.76 ({\bf actual 17})}
         \label{fig:subject3}
     \end{subfigure}
     \begin{subfigure}[b]{0.35\textwidth}
         \centering
         \includegraphics[width=\textwidth]{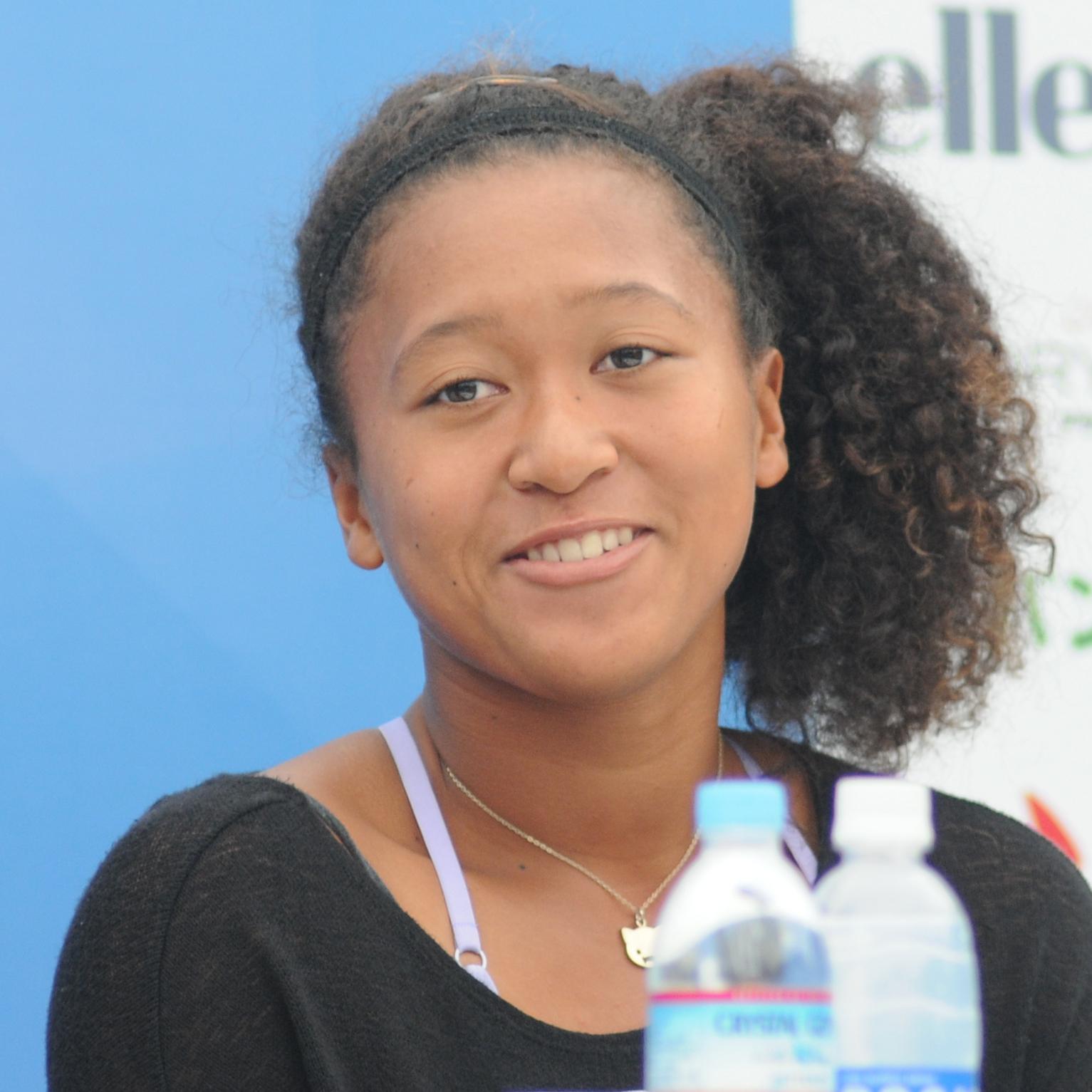}
         \caption{estimated 27.42 ({\bf actual 17})}
         \label{fig:subject4}
     \end{subfigure}
     \hfill
     \begin{subfigure}[b]{0.27\textwidth}
         \centering
         \includegraphics[width=0.83\textwidth]{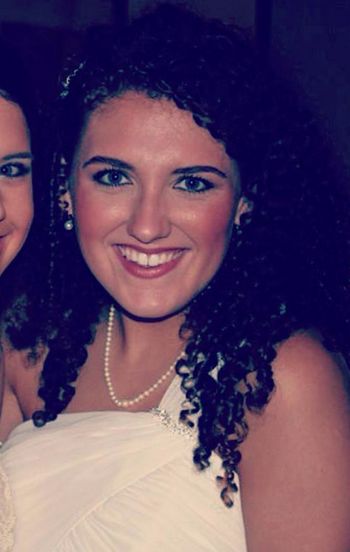}
         \caption{estimated 26.97 ({\bf actual 17})}
         \label{fig:subject5}
     \end{subfigure}
     \hfill
     \begin{subfigure}[b]{0.36\textwidth}
         \centering
         \includegraphics[width=0.97\textwidth]{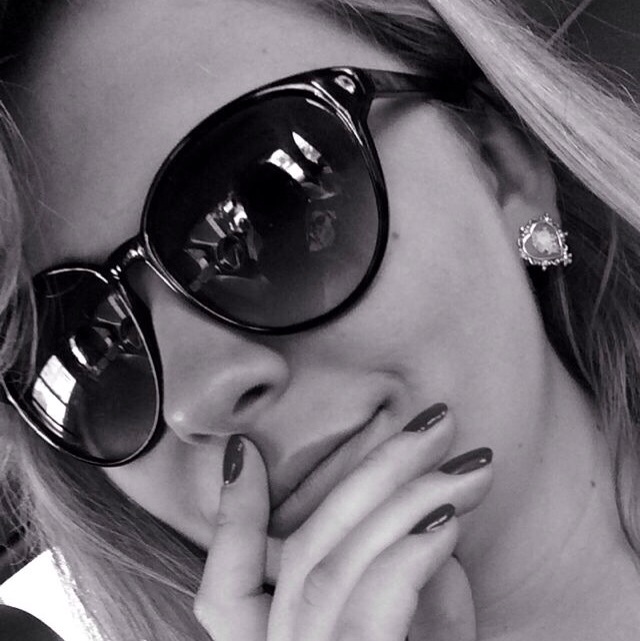}
         \caption{estimated 26 ({\bf actual 17})}
         \label{fig:subject6}
     \end{subfigure}
    \begin{subfigure}[b]{0.27\textwidth}
         \centering
         \includegraphics[width=\textwidth]{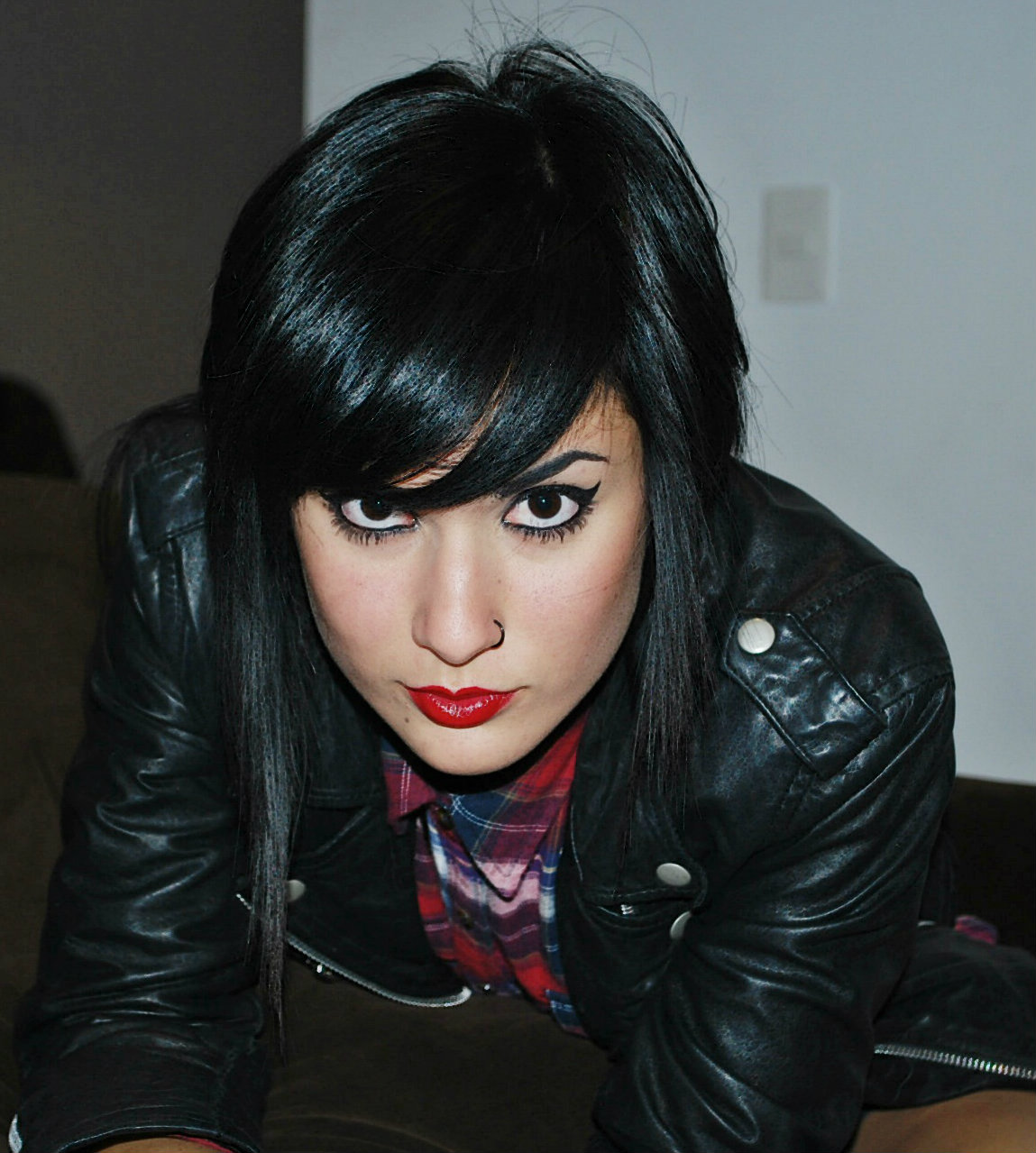}
         \caption{estimated 25.64 ({\bf actual 17})}
         \label{fig:subject7}
     \end{subfigure}
     \hfill
     \begin{subfigure}[b]{0.34\textwidth}
         \centering
         \includegraphics[width=0.89\textwidth]{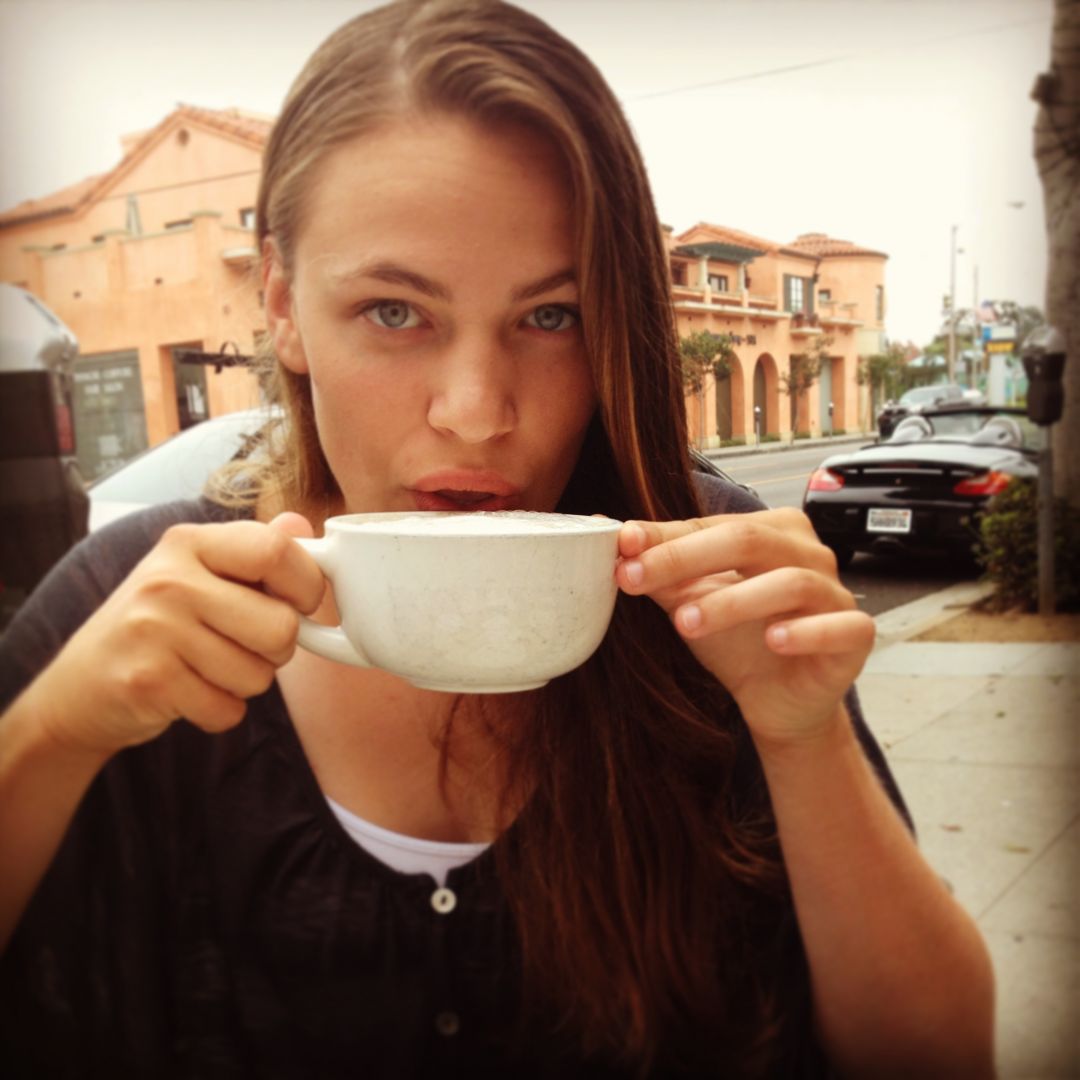}
         \caption{estimated 25.46 ({\bf actual 16})}
         \label{fig:subject8}
     \end{subfigure}
     \hfill
     \begin{subfigure}[b]{0.30\textwidth}
         \centering
         \includegraphics[width=\textwidth]{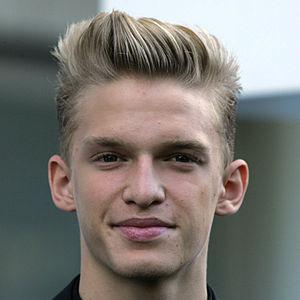}
         \caption{estimated 25.36 ({\bf actual 16})}
         \label{fig:subject9}
     \end{subfigure}     
        \caption{The nine subjects in the validation set of the appa-real database with actual age between 6 and 17 and average human estimated age higher than 25.}
        \label{fig:challengeData}
\end{figure*}

In the histogram in Figure~\ref{fig:appaReal-under18}, we first notice that, due to an outlier, a zero-failure test would require the operating threshold to be set at 43, that is, the ensemble of human estimators should operate a Challenge 43 policy. Apart from that single outlier, a zero-failure test on the other 214 subjects would require a Challenge 29 policy. In total, there were nine subjects estimated as being over 25. Figure~\ref{fig:challengeData} at the end of the paper shows these nine subjects together and their actual and estimated ages. Below are some observation on those images, which we believe are relevant to the design and construction of a set of positive samples for zero-failure testing: 
\begin{enumerate}
    \item The significant outlier with estimated age 42.81, shown in Figure~\ref{fig:subject1}, clearly had his actual age wrongly recorded as being just 6. This shows that in zero-failure testing there are smaller tolerances for this type of clerical errors compared to the computation of the usual performance metrics where such anomalies can be smoothed out. On the other hand, that could be seen as an advantage of zero-failure testing, as such data errors can go unnoticed with other testing methods while they still affect the quality of the assessment. 

    Regarding the other eight images in Figure~\ref{fig:challengeData}, any observations on them should be qualified by the caveat that there could be more, smaller and thus less obvious, errors in the reported actual ages. 

    \item In Figure~\ref{fig:subject2}, we have an image of unsatisfactory quality as it is grainy and half of the face has been cropped out. Figures~\ref{fig:subject3},~\ref{fig:subject5} might also have image quality issues by having been processed heavily. We believe that, in a well curated database for zero-failure testing, such lower quality images should be discarded, provided that we do not expect similar image quality issues under the actual deployment conditions of the system.

    \item Figures~\ref{fig:subject3},~\ref{fig:subject5},~\ref{fig:subject6},~\ref{fig:subject7} can be seen as {\em attack} samples, as, for example, the use of sunglasses, make-up, hair-style, or jewelry, may have increased or concealed the subject's apparent age. 

    We believe that a zero-failure test set should contain such attack samples as they can be expected under actual deployment conditions. However, following the discussion in Section~\ref{sec:sec3-2}, one has the option to tag them as such and create a separate attack samples test set $S_{att}$, and a separate stronger test consisting of the union of the regular and the attack test sets. That is, create the hierarchy of tests of increasing difficulty 
    $$
    S_{reg} \subset S_{reg} \cup S_{att}.
    $$

    \item Finally, Figures~\ref{fig:subject4},~\ref{fig:subject8},~\ref{fig:subject9} seem at a first glance as genuine failures of the estimators. They could give rise to some interesting comments on human perception and human biases, but such a discussion is out of the scope of the paper. 
\end{enumerate}

\subsubsection{Zero-failure test sets as requirement specifications and acceptance tests}

The above examples demonstrate that for a zero-failure test set to be useful in practice it should be highly curated, or otherwise it might be perceived as unnecessarily strict, or arbitrary and unfair. The construction a high-quality, professionally curated test dataset should be a systematic and transparent process. For example, one can conduct a human experiment, as the one in the appa-Real database, and if a sample is deemed too hard for humans to handle, it can be placed in a superset of the core test set and be used only for stricter zero-failure tests corresponding to higher levels of system certification. 

Adopting Software Engineering terminology, a highly curated zero-failure test set can be seen as an {\em acceptance test}, or if used earlier in the development cycle, as a {\em requirements specification}. We note that both acceptance tests and requirements specifications are critical parts of a system's development cycle, and thus, a considerable investment on them seems justified. 

Finally, we note that an approach to testing as the one we propose, would naturally lead to a sharper distinction between the desirable properties of the training and the test sets, which, in the current practice of machine learning, is rather mute. Indeed, it is very common to generate training and test sets by randomly sampling a single database, the only requirement being that the two sets are disjoint. Even when two different databases are used to generate the training and the test set, this is usually done for demonstrating the generalisation power of the classifier, while the choice of the training and the test database is considered interchangeable. 

To some extent, this practice is justified by the fact that high quality samples should work well both for training and for testing. However, such an argument overlooks the fact that quantity, the number of samples in a set, is more important in training than in testing. As a result, a database from which both a training and a test set will be sampled, tends to have orders of magnitude more samples than needed for robust testing, often at the expense of the quality of these test samples. In other words, the lack of a sharp distinction between the desirable properties of a training and a test set, means that the quantity required for robust training often comes at the expense of the quality required for robust testing.

\section{Conclusions}
\label{sec:sec4} 

We addressed the problem of assessing the performance of binary classifiers in asymmetric settings, where the costs of the two error types are incomparable. The metric we propose is the TNR when the algorithm's operating point is set to the lowest possible value that nevertheless gives zero failures on the set of positive samples. We illustrated the proposed with three age estimation examples, in the setting of determining whether a subject's age is under a legally defined threshold. We discussed several of the method's properties, and in particular, that it allows the construction of a hierarchy of tests of increasing difficulty. 

In the future, we would like to address, in the context of zero-failure testing, the problem of algorithmic bias across the various demographic characteristics of the subjects. In the application domain of presentation attack detection, \cite{idrd2022} and \cite{abduh2023} statistically analysed bias in type II errors. While the approaches there can also be applied here for the bias analysis of type II errors, the bias analysis of type I errors, within the proposed zero-failure testing framework, would probably require considerably different techniques.

\bibliographystyle{unsrt}
\bibliography{ref}

\clearpage


\end{document}